\documentclass{article}
\usepackage[dblblindworkshop, final]{neurips_2026}
\workshoptitle{The 6th Workshop on Mathematical Reasoning and AI}

\usepackage[utf8]{inputenc}
\usepackage[T1]{fontenc}
\usepackage{hyperref}
\usepackage{url}
\usepackage{booktabs}
\usepackage{amsfonts}
\usepackage{amsmath,amssymb}
\usepackage{microtype}
\usepackage{xcolor}
\usepackage{graphicx}

\title{When Is Proof-Step Verification Worth Its Cost?\\Operating Requirements of\\Certify-or-Decline Verification\\for Open-Weight Mathematical Reasoning\thanks{A version of this work also appears at the \emph{Who Verifies the Agents?} workshop at NeurIPS 2026.}}

\author{%
  {Abhinav Gupta}\\
  {Independent Researcher}
  \And
  {Ben Slivinski}\thanks{Equal contribution.}\\
  {Independent Researcher}
  \AND
  {Yashica Patodia}\footnotemark[2]\\
  {Independent Researcher}
  \And
  {Michael Saldivar}\\
  {Independent Researcher}
}

\begin{document}
\maketitle

\begin{abstract}
Proof-step verification rewrites a solver's answer into typed steps and uses LLM judges to certify or decline. We evaluate this pipeline on open-weight models across GPQA, AIME-2025, and a MATH-500 subset. Certification raises answer-key-match precision by 31 percentage points on the strong GPQA configuration and 46 on AIME, while answering only 35\% and 21\% of attempts. On near-ceiling MATH-500, gains are at most 5 points at a loss of 13--27 points of coverage. An output-token-matched self-consistency selector obtains 82.9\% precision at 35\% coverage, compared with the verifier's pooled 87.6\% (paired difference: $+4.8$ points, 95\% interval {$[-8.2,18.5]$)}. A diagnostic sample reveals a distinct failure mode: 52 of 88 completed problems end in invalid proof construction without ever reaching a judge. Finally, external-LLM acceptance of rendered certified chains is 61.0--67.6\%, compared with 87.6\% answer-key match. {These audits lack human or formal-kernel ground truth.} Open-weight auditor pairs show near-zero chance-corrected agreement, which measures consistency rather than accuracy against proof truth. These results motivate separate reporting of execution failures, selected-answer precision, and independently assessed proof validity. We provide per-configuration results and evaluation details in the supplement.
\end{abstract}

\section{Introduction}

Reliable mathematical reasoning needs more than a correct final answer: a supporting argument should establish its conclusion. Step-wise verification audits intermediate reasoning and retains answers whose steps are accepted \citep{ling2023deductive, lightman2023lets}. This induces a selective predictor \citep{geifman2017selective}, with precision and coverage that depend on the entire execution pipeline. We ask when certification improves selected-answer precision relative to answering everything, how it compares with self-consistency voting \citep{wang2023selfconsistency} under comparable token budgets, and whether a recorded decline reflects a judge's rejection or an earlier execution failure.

{We evaluate one existing architecture on open-weight stacks and separately audit accepted proof chains without revealing the answer key to the auditors. Certification selects answers with higher key-match precision, but the comparison with output-token-matched voting does not resolve an advantage. Most completed cases in a diagnostic run never reach a judge, and external auditors accept substantially fewer chains than the answer grader marks correct. These findings concern LLM-based filtering and its execution costs. Code and redacted run records are available at \url{https://github.com/yashica-patodia/certify-or-decline}.}

\section{System and measurement setup}

{The system is Theoria, introduced and evaluated with frontier models in our prior work \citep{theoria}.} A formalizer rewrites the solver's answer into typed reasoning-state transitions, each licensed by a citation, computation, or problem-given justification. Typed LLM judges audit steps in parallel, a pedantry filter screens rejections, and a bounded repair loop precedes certification or decline. Each role makes separate model calls within a bounded agent loop. {An attempt is \emph{certified} when the initial state and every step satisfy the pipeline's acceptance rules, including its filters. A \emph{convention lift} can resolve a rejection by recording an additional premise. Its prompt asks for a standard domain convention and a published source, but the convention's appropriateness and source are assessed by LLMs. Certification records acceptance of a structured informal argument under these rules, without a formal-kernel soundness guarantee.} {Supplement Section~\ref{sec:worked} gives a complete illustrative trace.}

Pipeline roles use Qwen3, gpt-oss-20b, and Phi-4 configurations \citep{qwen3, gptoss, phi4}. The strong configuration pairs Qwen3-14B judges with a gpt-oss-20b formalizer; the budget configuration uses Qwen3-14B. We evaluate 100 GPQA science questions \citep{gpqa}, all 30 AIME-2025 problems \citep{matharena}, and the first 100 MATH-500 rows \citep{hendrycks2021math, lightman2023lets}. {Initial-answer accuracy is 44--61\% on GPQA and AIME, leaving substantial headroom, while near-ceiling MATH supplies a contrasting regime.} Benchmark recency alone does not establish absence of training contamination.  {Repeated runs share problem sets and a fixed seed, chosen for reproducibility. Nondeterministic serving still makes runs differ, so we treat them as repeated measurements clustered by problem, not as independent-seed replications.} Coverage is certifications divided by all attempts, including execution errors; precision is key-matching certifications divided by certifications. {GPQA strong provides 105 certified rows from 47 distinct problems across three runs. Syntactic rules map 72 certified answers to an option, leaving 33 unmatched. On these 72 actual outputs, direct option scoring agrees with the answer grader in every case. The 33 free-form answers remain unchecked by this comparison. The grader scores verifier answers; self-consistency option letters are scored directly.} The supplement gives configurations, scoring checks, intervals, and audit procedures.

\section{Precision gains and coverage costs}

\begin{figure}[t]
\centering
\includegraphics[width=\textwidth]{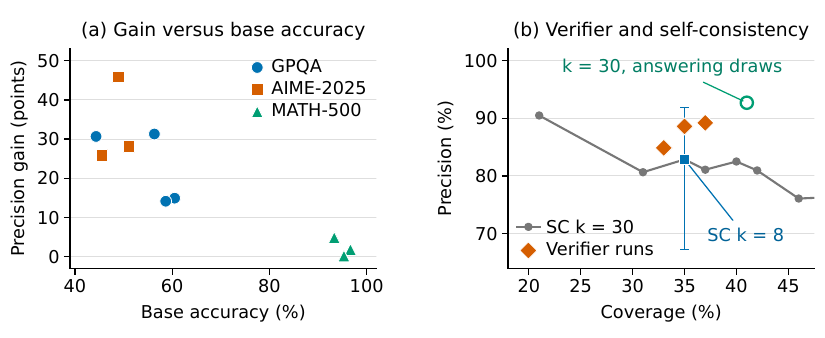}
\caption{(a) Precision gain versus initial-answer accuracy for Table~\ref{tab:value}. (b) GPQA verifier runs at 33\%, 35\%, and 37\% coverage; self-consistency (SC) $k=30$ with all draws in the agreement denominator (gray, low-coverage region); output-token-matched $k=8$ with a 95\% Wilson interval (blue); and $k=30$ with answering draws in the denominator (green). Lines connect measured thresholds as visual guides.}
\label{fig:value}
\end{figure}

Table~\ref{tab:value} compares certified precision with the initial solver draw, using the same answer grader. The largest gains occur on GPQA and AIME, while MATH is near ceiling (Figure~\ref{fig:value}a). This pattern is consistent with prior work on generator and verifier capability \citep{zhou2026variation}; here we quantify the associated loss of coverage. GPQA strong rises from 56.3\% initial-answer accuracy to 87.6\% certified precision while answering 35.0\% of attempts. AIME strong reaches 94.7\% on only 19 certifications across 90 attempts. MATH gains are 0.2--4.8 points, with 13.3--26.7 points of coverage lost. {These are selected-set gains, not additional problems answered correctly. Cross-benchmark differences confound difficulty with task and model, so we neither rank configurations nor infer a causal link between baseline accuracy and verification benefit.}

{The pooled GPQA strong figure combines 54 unconditional certifications and 51 accepted under conventions. Their key-match precision is 52/54 (96.3\%, cluster interval [90.6, 100.0]) and 40/51 (78.4\%, [62.5, 92.7]), respectively. They account for 18.0\% and 17.0\% of all attempts. These are different selected sets; the contrast does not estimate the effect of adding a convention.} {Supplement Section~\ref{sec:conventions} gives the breakdown and examples of recorded premises.}

\begin{table}[t]
\centering
\caption{Selected-answer precision and coverage. ``No verif.''\ is initial-answer accuracy over all problems; gain subtracts that baseline from certified precision. CIs use problem-cluster bootstrap. Small AIME samples and broad intervals preclude reliable configuration rankings. The GPQA budget baseline uses two graded runs (200 attempts), whereas verification pools four runs (400 attempts).}
\label{tab:value}
\begin{tabular}{llrrrr}
\toprule
Benchmark & Configuration & No verif. & Coverage & Precision [95\% CI] & Gain \\
\midrule
GPQA & strong & 56.3\% & 35.0\% & 87.6\% [77.7, 95.8] & \textbf{+31.3} \\
GPQA & budget & 60.5\% & 30.5\% & 75.4\% [64.1, 85.5] & +14.9 \\
GPQA & 32B-FP8 & 58.7\% & 45.3\% & 72.8\% [63.3, 81.6] & +14.1 \\
GPQA & Phi-4 & 44.3\% & 8.0\% & 75.0\% [52.6, 91.7] & +30.7 \\
AIME-2025 & strong & 48.9\% & 21.1\% & 94.7\% [77.8, 100.0] & \textbf{+45.8} \\
AIME-2025 & budget & 51.1\% & 26.7\% & 79.2\% [56.2, 95.8] & +28.1 \\
AIME-2025 & 32B-FP8 & 45.6\% & 31.1\% & 71.4\% [47.2, 92.0] & +25.9 \\
MATH-500 & strong & 96.7\% & 86.7\% & 98.5\% & +1.8 \\
MATH-500 & budget & 95.3\% & 74.0\% & 95.5\% & +0.2 \\
MATH-500 & 32B-FP8 & 93.3\% & 73.3\% & 98.2\% & +4.8 \\
\bottomrule
\end{tabular}
\end{table}

\section{Comparison with token-budget-matched sampling}

Prior work compares spending inference budgets on solving or verification \citep{singhi2025whentosolve}. {We make this comparison on GPQA because its option letters make votes exactly matchable. Voting over free-form AIME or MATH answers would first require clustering equivalent expressions, a matching step that adds complexity and its own ambiguity.} Our verifier averages 44{,}761 output and 71{,}628 input tokens per problem; a solver draw averages 5{,}284 and 795. Dividing these averages gives 8.47 output-matched or 19.15 total-token-matched draws, so we evaluate $k=8$ and $k=19$. Tokens are a budget proxy: heterogeneous models and input/output costs prevent equating this with FLOPs, latency, or monetary cost.

With non-answering draws counted as dissent, $k=8$ self-consistency selects 35/100 problems and answers 29 correctly, or 82.9\% precision (95\% Wilson interval 67.3--91.9\%). The verifier's pooled result is 92/105, or 87.6\%, at 35\% coverage; individual runs achieve 28/33, 31/35, and 33/37. A paired bootstrap over the same 100 problems, retaining all observed runs and draws per problem, gives a precision difference of $+4.8$ points (95\% interval {$[-8.2,18.5]$}; 20{,}000 resamples). {The interval includes zero, so the precision difference is not resolved at this sample size.} Under the answering-draw denominator, deterministic agreement thresholds attain minimum nonzero coverage of 57\% for $k=8$ and 47\% for $k=19$. At $k=30$, the most selective such point is 41\% coverage at 92.7\% precision, using 1.57 times the verifier's total-token budget. {These minima reflect finite threshold grids; randomized tie-breaking could lower them (Figure~\ref{fig:value}b). At the matched observed coverage, the data establish neither superiority nor equivalence. Voting with a larger budget reaches higher observed precision and coverage. The verifier also returns a step-audited chain, whose external assessment Section~\ref{sec:external} reports; we do not measure the utility of that artifact.}

\section{Construction failures hidden within declines}

Silent failures are a known agent-runtime hazard \citep{wu2026narratives}. A 100-problem diagnostic of the 14B default configuration had 12 execution errors and 88 completed cases. Of the latter, 60 ended in invalid formalization; 52 had never reached a judge, while eight had reached judging on an earlier attempt. Across the full diagnostic run, 117/277 formalizer responses placed a fact label in the answer slot. Thus a recorded decline cannot be interpreted uniformly as a judge rejecting an argument. Separately, 109/112 execution errors in GPQA strong and 65/69 in AIME strong were classified as agent-loop stalls. AIME strong produced 19 certifications, two recorded declines, and 69 execution errors across 90 attempts. Telemetry must distinguish construction failure, judge rejection, and execution error rather than treat all abstentions as evidence-sensitive decisions.

Across 1{,}878 rejections and 4{,}547 classified issues, the reported categories are hidden premises (44.4\%), misapplied theorems (24.6\%), fabricated citations (14.9\%), and arithmetic (11.2\%). Hidden premises, fabricated citations, and circular reasoning together account for 60.0\%. These are rejection-category frequencies, not independently validated detection rates. Without a holistic prose-review baseline, they do not establish that these issues are undetectable in prose.

\section{External assessment of certified proof chains}
\label{sec:external}

Answer matching and proof validity are distinct targets. External LLM auditors were supplied the problem, a rendering of the accepted chain, its conclusion, and stated conventions, but no answer key. {The rendering retains each step's final state entry and justification, while omitting complete intermediate state lists. Rejection can therefore reflect a reasoning defect, missing context, or auditor error.} One audited all 105 GPQA strong certifications; interrupted execution left a second audit with 57 completed rows. Across aggregation rules defined on all 105 rows, acceptance is 61.0--67.6\%, versus 87.6\% key-match precision, a 20--27-point discrepancy. {The full-coverage auditor labels 20 chains GAP (unsupported required steps) and 14 ERROR (incorrect steps).} In 25/105 rows, the answer matches the key while the full-coverage auditor rejects the rendered proof. These are \emph{LLM-assessed validity} rates, not human- or kernel-established proof soundness. On the 57-row overlap, both auditors accept 32 chains (56.1\%), while 49 answers match the key (86.0\%). Missing audits are not rejections; overlap results have a different cohort from the 105-row estimates.

The open-weight auditor pairs accept approximately 89--94\% of these already-certified chains and have near-zero Cohen's $\kappa$ (Table~\ref{tab:judgescale}). {Kappa measures chance-corrected consistency, not accuracy, and skewed acceptance rates depress it.} Known reasoning limitations and judge biases remain relevant \citep{zheng2023judging, norman2026reliability}. The cross-family judge swap gives 74.1\% precision at 19.3\% coverage, versus 75.4\% at 30.5\% for budget; different accepted sets and absent uncertainty estimates prevent using this swap to rule out preference leakage \citep{preferenceleakage}.

\begin{table}[t]
\centering
\caption{Inter-auditor agreement on certified GPQA chains. Overlap sizes differ. These are consistency statistics, not accuracy against independently adjudicated proof validity.}
\label{tab:judgescale}
\begin{tabular}{lcc}
\toprule
Judge pair & Overlap & Cohen's $\kappa$ \\
\midrule
Proprietary (Claude-family + GPT-family) & 57 & 0.573 \\
gpt-oss-20b + Qwen3-14B & 105 & 0.038 \\
Qwen3-32B + gpt-oss-20b & 105 & $-0.005$ \\
\bottomrule
\end{tabular}
\end{table}

\section{Operating requirements and limitations}

Deployment should account for three separate quantities: selected-answer precision, coverage lost to failed execution, and evidence that accepted reasoning is valid. High precision at low coverage can be useful when incorrect answers are costly, but this study specifies no application utility function. Prompt variants have observed precision differences of 7--11 points relative to budget, with broad intervals; this motivates robustness checks without establishing a universal effect-size floor. Construction and repair budgets likewise need explicit reporting.

The study covers one architecture and a small set of model configurations. GPQA is science QA; AIME has only 30 problems; MATH is near ceiling. The validity audit lacks human adjudication, and the partial second-auditor cohort can introduce selection effects. {No controlled run replaces the internal judges with frontier models. The prior frontier evaluation uses different cohorts and settings, so it cannot explain the present audit gap. The mechanical scoring check leaves the 33 free-form answers unvalidated; independent annotation is needed to assess the scoring difference between selectors.} A token-matched holistic single-call judge was not evaluated. Human-annotated step-level benchmarks such as Hard2Verify \citep{hard2verify} provide complementary evaluations. Independently seeded replications and independent proof annotation are needed to resolve comparative performance and actual soundness.

{A formalized subset checked by Lean \citep{lean4} would provide an independent reference for evaluating internal judges and external auditors. Its formal statements must also be checked against the original problems. Computer algebra could separately check computation steps, but does not supply a proof-kernel guarantee for an entire informal chain. Judges in this study ran with shell and CAS tools disabled.}

\section{Conclusion}

This open-weight case study finds substantial precision--coverage tradeoffs, construction failures hidden within declines, and a gap between answer-key match and external-LLM proof acceptance. {The experiments do not establish an answer-selection advantage over output-token-matched voting. Theoria also returns an inspectable chain, but the external audit exposes unresolved reliability concerns. Whether that artifact warrants its cost requires an application-specific evaluation with independently validated reasoning.}

\clearpage
\appendix
\setcounter{table}{0}
\renewcommand{\thetable}{S\arabic{table}}
\renewcommand{\theHtable}{S\arabic{table}}
\def\mathaicombined{1}
\ifdefined\mathaicombined
\else
\documentclass{article}
\usepackage[dblblindworkshop, final]{neurips_2026}
\workshoptitle{The 6th Workshop on Mathematical Reasoning and AI}
\usepackage[utf8]{inputenc}
\usepackage[T1]{fontenc}
\usepackage{booktabs}
\usepackage{graphicx}
\usepackage{amsmath,amssymb}
\usepackage{microtype}
\usepackage{url}
\usepackage{xcolor}
\renewcommand{\thetable}{S\arabic{table}}
\title{When Is Proof-Step Verification Worth Its Cost?\\Operating Requirements of\\Certify-or-Decline Verification\\for Open-Weight Mathematical Reasoning}
\begin{document}
\maketitle
\fi

\raggedbottom
\ifdefined\mathaicombined
\section*{Supplementary material}
\fi

This supplement reports the per-arm results, coverage decomposition,
self-consistency operating points, rejection labels, and grader checks behind
the main text. {Code is available at \url{https://github.com/yashica-patodia/certify-or-decline}.}

\section{Measurement definitions and per-arm results}

The fixed GPQA set contains 100 problems from GPQA Diamond: an initial
25-problem set plus a frozen 75-problem extension. AIME uses all 30 problems
from the 2025 AIME I/II dataset. MATH uses the first 100 test rows of
MATH-500 in dataset order, so its subset is not a random sample.

The strong configuration uses Qwen3-14B for the solver, typed judges, pedantry
filter and convention lift, with gpt-oss-20b as formalizer. The budget
configuration uses Qwen3-14B throughout. The 32B and Phi-4 arms change the whole
stack to Qwen3-32B-FP8 and Phi-4, respectively. The strong, budget, 32B, AIME
and MATH call settings use temperature 0.6, top-$p$ 0.95, an 8{,}192-token
output cap, a 32{,}768-token context and a 16-turn role limit, with search and
shell tools disabled; Phi-4 uses a 16{,}384-token context. Recorded calls reuse
seed 20260718, so repeated runs are not distinct-seed replications. Run-to-run
variation is retained, without assuming independent random seeds.

All configurations allow up to three solver answers. The \emph{ctl} arm uses
Qwen3-14B throughout, with ceilings of three verification attempts and three
invalid-formalization attempts; \emph{fswap} replaces only its formalizer with
gpt-oss-20b. The budget, strong, 32B, Phi-4, crossjudge, prompt A/B, AIME, and
MATH arms raise these ceilings to six and ten, respectively. The
\emph{fswap+budget} row is a separate run group with the same model-role
assignment and ceilings as strong.

A run attempt is one problem evaluated in one repeated run. With $P$ problems
and $k$ runs, there are $N=kP$ attempts; a problem certified in multiple runs
contributes multiple certified rows. If $C$ attempts are certified and $Q$ of
those answers match the key under the answer grader, coverage is $C/N$ and
certified precision is $Q/C$. Certification pools unconditional acceptance under the pipeline's rules and acceptance under recorded conventions. In the strong GPQA arm these comprise 54 and 51 of the 105 certified rows, respectively. Execution errors remain in $N$ and therefore reduce coverage. The no-verification baseline grades the solver's first draw
over all problems. For GPQA budget, the initial-answer baseline has 121/200 key matches from two graded runs, while its verifier result pools four runs (400 attempts). The other headline baselines use the reported three runs. These gains compare pooled summaries, rather than paired before/after outcomes. Key-match precision measures answer agreement with the key; the separate chain-validity audit below measures an external auditor's judgment
of the accepted reasoning.

Table~\ref{tab:arms} reports the eleven arms with cluster-bootstrap intervals.
Problems are resampled with all their runs attached, using 20{,}000 resamples.
These intervals account for repeated observations of the same problem; they do not
account for systematic grading error. Table~\ref{tab:seeds} reports counts and
per-run precision. The prompt A/B rows and other configuration changes are
descriptive comparisons; their differences do not isolate prompt effects from
sampling variation.

\begin{table}[htbp]
\centering
\caption{Per-arm coverage and certified precision, with 95\% cluster-bootstrap
intervals over problems.}
\label{tab:arms}
\begin{tabular}{@{}lrr@{}}
\toprule
Arm & Coverage [95\% CI] & Precision [95\% CI] \\
\midrule
GPQA strong & 35.0\% [27.0, 43.3] & 87.6\% [77.7, 95.8] \\
GPQA budget & 30.5\% [24.5, 36.5] & 75.4\% [64.1, 85.5] \\
GPQA fswap & 30.0\% [23.3, 37.0] & 90.0\% [81.9, 96.2] \\
GPQA 32B & 45.3\% [38.7, 52.0] & 72.8\% [63.3, 81.6] \\
GPQA Phi-4 & 8.0\% [4.3, 12.0] & 75.0\% [52.6, 91.7] \\
GPQA effort=high & 9.0\% [4.0, 15.0] & 88.9\% [62.5, 100.0] \\
GPQA judge-prompt A & 35.5\% [28.0, 43.5] & 64.8\% [50.6, 78.3] \\
GPQA judge-prompt B & 30.0\% [24.0, 36.0] & 67.8\% [55.4, 79.2] \\
AIME strong & 21.1\% [8.9, 35.6] & 94.7\% [77.8, 100.0] \\
AIME budget & 26.7\% [14.4, 40.0] & 79.2\% [56.2, 95.8] \\
AIME 32B & 31.1\% [18.9, 43.3] & 71.4\% [47.2, 92.0] \\
\bottomrule
\end{tabular}
\end{table}

\begin{table}[htbp]
\centering
\setlength{\tabcolsep}{3pt}
\caption{Sample counts and variation across runs. $P$ counts problems and $C$
counts certified problem--run rows. Per-run precision is in percent; spread is
the reported range in percentage points. The effort=high arm has one run.}
\label{tab:seeds}
\begin{tabular}{@{}p{3.5cm}rrrp{3.5cm}r@{}}
\toprule
Arm & $k$ & $P$ & $C$ & Per-run precision & Spread \\
\midrule
GPQA strong & 3 & 100 & 105 & 84.8 / 88.6 / 89.2 & 4.4 \\
GPQA budget & 4 & 100 & 122 & 71.0 / 82.8 / 67.9 / 79.4 & 14.9 \\
GPQA fswap & 3 & 100 & 90 & 84.8 / 93.1 / 92.9 & 8.3 \\
GPQA 32B & 3 & 100 & 136 & 82.0 / 65.9 / 68.9 & 16.1 \\
GPQA Phi-4 & 3 & 100 & 24 & 63.6 / 83.3 / 85.7 & 22.1 \\
GPQA effort=high & 1 & 100 & 9 & 88.9 & -- \\
GPQA judge-prompt A & 2 & 100 & 71 & 67.6 / 62.2 & 5.4 \\
GPQA judge-prompt B & 3 & 100 & 90 & 65.5 / 70.6 / 66.7 & 5.1 \\
AIME strong & 3 & 30 & 19 & 100.0 / 100.0 / 85.7 & 14.3 \\
AIME budget & 3 & 30 & 24 & 66.7 / 80.0 / 87.5 & 20.8 \\
AIME 32B & 3 & 30 & 28 & 70.0 / 85.7 / 63.6 & 22.1 \\
\bottomrule
\end{tabular}
\end{table}

Table~\ref{tab:points} reports additional arms as point estimates; no cluster
interval was computed for these. The \emph{crossjudge} arm changes the judges
to gpt-oss-20b. Its 74.1\% precision is 1.3 percentage points below the budget
arm's 75.4\%. This comparison does not establish equivalence or rule out
shared model biases. It differs from the Phi-4 arm, which changes the whole
stack rather than only the judges. The \emph{effort} arms change formalizer reasoning effort.

\begin{table}[htbp]
\centering
\setlength{\tabcolsep}{4pt}
\caption{Additional arms reported without intervals. $N$ is the number of
problem--run attempts.}
\label{tab:points}
\begin{tabular}{@{}llrrrr@{}}
\toprule
Benchmark & Arm & $k$ & $N$ & Coverage & Precision \\
\midrule
GPQA & ctl & 4 & 400 & 14.5\% & 81.0\% \\
GPQA & fswap+budget & 3 & 300 & 36.0\% & 86.1\% \\
GPQA & crossjudge & 3 & 300 & 19.3\% & 74.1\% \\
GPQA & minus-completeness & 3 & 300 & 35.7\% & 72.0\% \\
GPQA & effort=low & 3 & 300 & 11.0\% & 81.8\% \\
MATH-500 & strong & 3 & 300 & 86.7\% & 98.5\% \\
MATH-500 & budget & 3 & 300 & 74.0\% & 95.5\% \\
MATH-500 & 32B & 3 & 300 & 73.3\% & 98.2\% \\
\bottomrule
\end{tabular}
\end{table}

\section{Unconditional and convention-dependent certification}
\label{sec:conventions}

Table~\ref{tab:tiers} separates the two certification types in GPQA strong. The intervals resample all 100 problem clusters, retaining their three runs, with 20{,}000 draws and seed 20260819. Problems can contribute rows to both types across runs. Their different selected sets prevent interpreting the precision contrast as the causal effect of convention lifting.

\begin{table}[htbp]
\centering
\caption{GPQA strong certification types. Coverage uses all 300 attempts; precision uses the certified rows of the indicated type.}
\label{tab:tiers}
\begin{tabular}{@{}lrrr@{}}
\toprule
Acceptance type & Key matches / certified & Coverage & Precision [95\% CI] \\
\midrule
Unconditional & 52 / 54 & 18.0\% & 96.3\% [90.6, 100.0] \\
Under recorded conventions & 40 / 51 & 17.0\% & 78.4\% [62.5, 92.7] \\
Pooled & 92 / 105 & 35.0\% & 87.6\% [77.7, 95.8] \\
\bottomrule
\end{tabular}
\end{table}

Recorded premises include using natural units ($c=1$) in a relativistic calculation and restricting a discussion of high-energy regularization to quantum field theories. The first fixes units; the second narrows the scope of the question. A chain may record several lifts at different steps. These are examples of premises accepted by the system, rather than independently verified conventions. The lift model supplies its own source attribution, which the experiment does not independently validate.

\section{Coverage and recorded termination labels}

Table~\ref{tab:decomp} decomposes coverage using the pipeline's recorded labels.
\emph{Answered} denotes certification. A recorded \emph{decline} can follow
judge rejection or failure to construct an acceptable proof before any judge
runs; it is not necessarily a verifier judgment. \emph{Execution error}
denotes failure to finish the agent loop. These categories sum to the number
of attempts, but the decline column does not separate its two causes.
Agent-loop stalls dominate the execution errors in the reported runs.

The construction diagnostic is a separate 100-problem run with 12 execution
errors and 88 completed cases. Of those 88, 60 end in invalid formalization,
but eight had reached judging on an earlier attempt; 52 had never reached a
judge. Another 15 completed cases end in formalizer rejection without judging,
so 67/88 never reach a judge overall. The first formalizer decisions are 64
invalid outputs and 24 rejections, with none proceeding to judging. Across all
100 problems, 117 of 277 recorded proof responses place a fact label in the
answer slot; restricting to the 88 completed cases gives 109/264. These
diagnostic denominators differ from the intention-to-treat coverage tables.

The existing formalizer role swap offers a limited probe of this bottleneck. With the default retry ceilings held fixed, the Qwen3-14B control has 58/400 certifications (14.5\% coverage), versus 90/300 (30.0\%) when only the formalizer is changed to gpt-oss-20b. These arms reuse the same problems but have different solver draws and run counts. Construction failures were not separately traced across both arms, so the coverage difference does not identify whether model capacity, schema compliance, or another pipeline interaction caused the improvement.

\begin{table}[htbp]
\centering
\setlength{\tabcolsep}{4pt}
\caption{Coverage decomposed by recorded termination label. Execution-error
percentages use all attempts as the denominator.}
\label{tab:decomp}
\begin{tabular}{@{}lrrrr@{}}
\toprule
Arm & Attempts & Answered & Declined & Execution error \\
\midrule
GPQA strong & 300 & 105 & 83 & 112 (37\%) \\
GPQA budget & 400 & 122 & 182 & 96 (24\%) \\
GPQA fswap & 300 & 90 & 111 & 99 (33\%) \\
GPQA 32B & 300 & 136 & 61 & 103 (34\%) \\
GPQA Phi-4 & 300 & 24 & 241 & 35 (12\%) \\
GPQA effort=high & 100 & 9 & 10 & 81 (81\%) \\
GPQA judge-prompt A & 200 & 71 & 89 & 40 (20\%) \\
GPQA judge-prompt B & 300 & 90 & 139 & 71 (24\%) \\
AIME strong & 90 & 19 & 2 & 69 (77\%) \\
AIME budget & 90 & 24 & 10 & 56 (62\%) \\
AIME 32B & 90 & 28 & 6 & 56 (62\%) \\
\bottomrule
\end{tabular}
\end{table}

\section{Self-consistency under two vote conventions}

The token-budget comparison uses $k=8$ solver draws for an output-token match
and $k=19$ for an input-plus-output-token match; $k=30$ supplies a higher-budget
curve. These budgets use nested prefixes of the same 30-draw corpus, so
cross-budget agreement is not independent replication. Here $k$ counts
solver draws per problem, unlike the run count in the
preceding tables. Token counts are compute proxies, not measured FLOPs or
latency. The verifier averages 44{,}761 output and 71{,}628 input tokens per
problem, versus 5{,}284 output and 795 input tokens for one solver draw.

{Self-consistency sampling uses Qwen3-14B with temperature 0.6, top-$p$ 0.95, top-$k$ 20, and a maximum output length of 8{,}192 tokens. Draw $i$ uses seed $20260718+1000i$, for $i=0,\ldots,29$. The verifier configurations do not explicitly set top-$k$.}

Let $m$ be the largest option-letter vote count and $a$ the number of draws
with a parsed answer. \emph{Answering draws} uses agreement $m/a$ (when $a>0$);
\emph{all draws} uses $m/k$, so non-answering draws count as dissent. Applying
agreement thresholds produces the operating points in Table~\ref{tab:curves}.
The table lists observed threshold points: failure to reach exactly 35\%
coverage refers to these discrete thresholds, and does not preclude randomized
tie-breaking or subsampling. No interpolation is used. Table~\ref{tab:draws}
shows the parsing counts behind the denominator difference. Answering-draw
unanimity imposes no minimum response count: two of its 41 selected problems
at $k=30$ have only three and four parsed draws. Requiring at least five
parsed draws yields 37/39 correct (94.9\%) at 39\% coverage, versus 38/41
(92.7\%) without that requirement.

\begin{table}[htbp]
\centering
\setlength{\tabcolsep}{3pt}
\caption{Observed threshold operating points. Pairs are coverage / precision,
in percent. The fourth column gives precision at exactly 35\% coverage;
``--'' means no reported threshold reaches that coverage.}
\label{tab:curves}
\begin{tabular}{@{}rlrrr@{}}
\toprule
$k$ & Vote denominator & Most selective & At 35\% & Maximum coverage \\
\midrule
8 & Answering draws & 57.0 / 82.5 & -- & 99.0 / 63.6 \\
8 & All draws & 35.0 / 82.9 & 82.9 (29/35) & 99.0 / 63.6 \\
19 & Answering draws & 47.0 / 85.1 & -- & 100.0 / 65.0 \\
19 & All draws & 28.0 / 82.1 & -- & 100.0 / 65.0 \\
30 & Answering draws & 41.0 / 92.7 & -- & 100.0 / 63.0 \\
30 & All draws & 21.0 / 90.5 & 82.9 (29/35) & 100.0 / 63.0 \\
\bottomrule
\end{tabular}
\end{table}

\begin{table}[htbp]
\centering
\caption{Draw parsing. Empty completions account for most unparsed draws.}
\label{tab:draws}
\begin{tabular}{@{}rrrrr@{}}
\toprule
$k$ & Total draws & Parsed & Unparsed & Of these, empty \\
\midrule
8 & 800 & 693 & 107 & 98 \\
19 & 1,900 & 1,635 & 265 & 229 \\
30 & 3,000 & 2,576 & 424 & 358 \\
\bottomrule
\end{tabular}
\end{table}

Both selectors are evaluated on the same problems. The baseline's votes and
the answer key are option letters and are scored directly. The verifier's
free-form answers use the model grader checked in Section~\ref{sec:grader}.
Thus the two arms have different scoring mechanisms, and the constructed-case
checks do not establish their equivalence on actual model outputs. An earlier
comparison used letter matching for the verifier's free-form answers; the
results here use the corrected scoring procedure.

For the $k=8$ all-draws comparison, the observed policy requires all eight
draws to produce the same parsed option. It accepts 35 problems, answering 29
correctly. The three verifier runs certify 33, 35 and 37 problems, with 28,
31 and 33 correct answers, respectively. Pooled verifier precision minus
self-consistency precision is $92/105-29/35=0.04762$.

An exploratory paired bootstrap uses 20{,}000 resamples of the 100 problem IDs
with replacement (seed 20260906). Each sampled problem carries all three
verifier runs and eight self-consistency draws. Each method retains its fixed
selection policy; we recompute both precision ratios on their own accepted
sets, without restricting to their intersection or retuning coverage to 35\%.
All denominators are defined. The 2.5th and 97.5th percentiles give
$[-8.2,18.5]$ percentage points. The released calculation uses Python \texttt{random.Random} with the problem IDs sorted before sampling. This interval conditions on observed runs, draws and grader labels, excluding new-run randomness and systematic grading
error. Resampled coverages can differ despite both observed coverages being
35\%. With zero inside the interval and no prespecified equivalence margin,
neither superiority nor equivalence is established.

\section{Rejection composition}

Table~\ref{tab:reject} reports assigned error-class labels, rather than
independently established errors. Hidden premises, fabricated citations, and
circular reasoning together account for 2{,}728 of 4{,}547 issues (60.0\%).
This composition does not establish which issues a prose-based or human
reviewer could detect; no such comparison is reported in this campaign.

\begin{table}[htbp]
\centering
\caption{Assigned issue labels across 1{,}878 rejections and 4{,}547 issues
from 67 run files. Multiple issues can be associated with one rejection.}
\label{tab:reject}
\begin{tabular}{@{}lrr@{}}
\toprule
Error class & Issues & Share \\
\midrule
Hidden premise & 2020 & 44.4\% \\
Misapplied theorem & 1117 & 24.6\% \\
Fabricated citation & 678 & 14.9\% \\
Arithmetic error & 510 & 11.2\% \\
Other & 163 & 3.6\% \\
Circular reasoning & 30 & 0.7\% \\
Cosmetic & 29 & 0.6\% \\
\bottomrule
\end{tabular}
\end{table}

By judge role, computation accounts for 44.2\% of rejections, citation
37.9\%, problem-given 17.5\%, and untyped steps the remaining 0.4\%.

\section{Model-assessed validity of certified chains}

Table~\ref{tab:sound} reports four aggregation rules for the chain-validity
audit. The GPT-family auditor, labeled \emph{Codex} in the audit, evaluated
all 105 certified problem--run rows of the strong GPQA arm, representing 47
distinct problems. The Claude-family auditor evaluated 57 of those rows. Its
audit was incomplete because the foreground-only grading process was resumed
in chunks; the other 48 entries are missing verdicts, not rejections.

The configured proprietary auditors are \texttt{gpt-5.6-terra} with
\texttt{xhigh} reasoning effort and the Claude model alias \texttt{opus} with
\texttt{max} effort. The latter alias does not pin a resolved model version.
The GPT-family audit configuration enables search. The open-weight auditors
are \texttt{openai/gpt-oss-20b}, \texttt{Qwen/Qwen3-14B}, and
\texttt{Qwen/Qwen3-32B-FP8}, with temperature 0, seed 20260718, a 4{,}096-token
output cap, 32{,}768-token context, and an eight-turn limit. Their search and
shell tools are disabled; gpt-oss uses low reasoning effort. These model and
tool settings differ, so the pair comparisons do not isolate parameter count.

Auditors receive the problem, initial state, each step's final state entry and
justification, the extracted conclusion, and any accepted conventions. The
answer key is not supplied in the audit prompt, and the rendering omits the
complete intermediate state lists. The
rubric asks whether the rendered chain establishes its conclusion without
solving the problem independently; standard facts may be asserted, but the
problem's crux requires support. Verdicts are VALID, GAP (unsupported required
step), ERROR (incorrect step), or NONSEQUITUR (conclusion not established),
with the first defective step and a short reason. Thus this is an audit of
rendered chains; omitted state entries may affect the judgment. No human or
formal-kernel ground-truth audit is provided here.

The full-coverage auditor assigns 20 GAP and 14 ERROR verdicts, accounting for all 34 rejected chains; it assigns no NONSEQUITUR verdict. GAP denotes a required claim without support, whereas ERROR denotes an incorrect claim or calculation. These labels describe the auditor's reasons, not independently adjudicated error classes. They also do not distinguish an omitted state entry from a gap in the underlying complete proof.

The three rules using all 105 rows give acceptance rates of 61.0--67.6\%,
20.0--26.6 percentage points below key-match precision on those rows
(87.6\%). The both-VALID rule uses only the 57-row overlap; its 56.1\% rate
cannot be compared directly with the 105-row key-match rate. The mixed rules
also depend on the operationally incomplete second audit. Within the 57-row
overlap, key-match precision is 49/57 (86.0\%), Codex acceptance is 39/57
(68.4\%), and Claude acceptance is 36/57 (63.2\%). Of the 105 rows, 25 have a
key-matching answer on a chain rejected by the full-coverage auditor; this
count rises to 31 when either auditor's rejection is used.

\begin{table}[htbp]
\centering
\setlength{\tabcolsep}{4pt}
\caption{Auditor acceptance by aggregation rule. The denominator is explicit
because the final row uses a different cohort.}
\label{tab:sound}
\begin{tabular}{@{}p{3.1cm}rrp{5.1cm}@{}}
\toprule
Rule & Accepted & Rate & Definition \\
\midrule
Codex only & 71 / 105 & 67.6\% & Use the auditor with full coverage. \\
\addlinespace
Claude-else-Codex & 68 / 105 & 64.8\% & Use Claude where available, otherwise Codex. \\
\addlinespace
Neither auditor rejects & 64 / 105 & 61.0\% & Retain rows unless an available auditor rejects them. \\
\addlinespace
Both auditors VALID & 32 / 57 & 56.1\% & Require both auditors to accept, on the overlap only. \\
\bottomrule
\end{tabular}
\end{table}

On the 57-row overlap, both proprietary auditors accept 32 rows and both
reject 14; Codex alone accepts seven and Claude alone accepts four. Collapsing
GAP, ERROR, and NONSEQUITUR to rejection gives Cohen's $\kappa=0.573$. The open-weight pairs in the main text audit all 105 rows.
Cohen's $\kappa$ measures agreement beyond marginal label frequencies, not
accuracy against proof truth. High acceptance on this certified cohort and
near-zero $\kappa$ establish neither the cause of disagreement nor general
proof-auditing ability.

\section{Answer-grader checks on constructed cases}
\label{sec:grader}

The answer grader is \texttt{openai/gpt-oss-20b} in native MXFP4 precision,
with low reasoning effort, temperature 0, seed 20260718, a 4{,}096-token output
cap, 32{,}768-token context, and a two-turn limit. Search and shell tools are
disabled. It sees the problem, canonical answer and selected candidate, with
experimental condition and verification status hidden. Its rubric accepts
exact, algebraic, notational, or semantic equivalence; it rejects placeholders,
refusals, unresolved symbols, incompatible alternatives, and incomplete answers.
A correct argument elsewhere cannot rescue a wrong selected answer. This
key-matching task differs from the key-blind chain audit above.

Table~\ref{tab:grader} tests this grader with the same prompt, schema, and call
path used for run scoring. GPQA supplies 40 cases each of correct/incorrect
option letters, correct/incorrect option text, and empty answers. AIME uses
30 each of exact answers, different answers, off-by-one integers, trailing
punctuation, and empty answers, plus nine zero-padding variants. MATH uses
30 each of exact, different, and empty answers, 14 notation variants, four
whitespace variants, and six off-by-one integers.

\begin{table}[htbp]
\centering
\setlength{\tabcolsep}{4pt}
\caption{Answer-grader checks. False positives accept incorrect or empty
answers; false negatives reject correct answers. No calls failed.}
\label{tab:grader}
\begin{tabular}{@{}lrrrr@{}}
\toprule
Benchmark & Cases & Accuracy & False positives & False negatives \\
\midrule
GPQA & 200 & 100.0\% & 0 & 0 \\
AIME-2025 & 159 & 98.7\% & 2 & 0 \\
MATH-500 & 114 & 99.1\% & 1 & 0 \\
\bottomrule
\end{tabular}
\end{table}

One answer-grader call on an actual certified MATH budget output fails and is scored incorrect in the aggregate; it is separate from the constructed-case checks above.

All three errors concern empty answers (28/30 AIME and 29/30 MATH empty cases
correct). Certified records have no empty final answers; the loader marks empty answers incorrect. For an additional check on generated outputs, we applied the existing syntactic option rules to all 105 GPQA strong certifications. Sixty name an option letter and twelve reproduce one option's text. Direct key matching agrees with the LLM grader on all 72 such outputs. The remaining 33 cannot be resolved by those rules. This mechanical check uses actual outputs but provides no independent annotation of free-form answers, paraphrases, or derived quantities. The numerical comparison is published as \texttt{scoring\_consistency.json} in the repository.

\clearpage

\section{Worked example of the recorded proof format}
\label{sec:worked}

The repository includes a complete trace for ``Prove there are infinitely many primes,'' taken from the public example accompanying the prior Theoria work. It illustrates the format and acceptance rules; it is not an observation from the open-weight benchmark experiments in this paper. Table~\ref{tab:worked} gives every transition, with notation shortened for readability. The complete state lists and recorded justifications are supplied in \texttt{examples/euclid\_primes.json} in the paper repository.

The initial state contains only the goal. Through step 15, each state retains its predecessor's entries and adds the assertion listed below, except that step 2 also shortens the label ``Assumption $H$ (for contradiction)'' to ``Assumption $H$.'' Step 16 replaces the whole state with the conclusion after discharging $H$. Thus these transition rules specify every state, including the temporary assumption's scope.

\begin{table}[htbp]
\centering
\small
\caption{All sixteen transitions of the recorded Euclid example. C denotes citation and M computation; these types select LLM judge roles.}
\label{tab:worked}
\begin{tabular}{@{}rcp{4.0cm}p{4.8cm}@{}}
\toprule
Step & Type & Assertion added or final state & Recorded justification, paraphrased \\
\midrule
1 & C & $H$: only finitely many primes exist. & Open a proof by contradiction. \\
2 & C & $2$ is prime; shorten $H$'s label. & Definition of a prime. \\
3 & C & $p_1,\ldots,p_n$ list all primes, $n\geq1$. & A finite nonempty set can be enumerated. \\
4 & C & Each $p_i\geq2$. & Every listed number is prime. \\
5 & M & $N:=p_1\cdots p_n+1$. & Introduce a fresh integer by definition. \\
6 & M & $N>1$. & The nonempty product is at least $2$. \\
7 & C & Every integer greater than $1$ has a prime divisor. & Its least divisor greater than $1$ is prime. \\
8 & C & Some prime $q$ divides $N$. & Apply step 7 to $N>1$. \\
9 & C & Each $p_i$ divides $p_1\cdots p_n$. & A factor divides its product. \\
10 & C & $d\mid a$ and $d\mid b$ imply $d\mid(a-b)$. & Subtract integer multiples of $d$. \\
11 & M & $p_i\mid N$ implies $p_i\mid1$. & Subtract the product from $N$. \\
12 & C & No prime divides $1$. & A prime is at least $2$. \\
13 & C & No $p_i$ divides $N$. & Steps 11 and 12, by modus tollens. \\
14 & C & $q\ne p_i$ for every $i$. & Otherwise $p_i$ divides $N$, contrary to step 13. \\
15 & C & Contradiction. & $q$ is prime but absent from the complete list. \\
16 & C & Final state: infinitely many primes exist. & Discharge $H$ by contradiction. \\
\bottomrule
\end{tabular}
\end{table}

The initial-state audit and all final step verdicts are accepted. At step 2, the citation judge initially rejects the shortening of $H$'s label; the pedantry filter overrides this as a notational change. The trace is therefore recorded as unconditionally certified, with no convention lifts, even though one raw judge rejection was overridden. This example shows why certification must be defined through the pipeline's acceptance rules. No formal proof kernel checks this trace.

\ifdefined\mathaicombined
\else
\end{document}
\fi


\clearpage
\begin{thebibliography}{19}

{\bibitem[de~Moura and Ullrich(2021)]{lean4}
L.~de~Moura and S.~Ullrich.
\newblock The {Lean 4} theorem prover and programming language.
\newblock In \emph{CADE}, 2021.}

\bibitem[Saldivar and Slivinski(2026)]{theoria}
M.~Saldivar and B.~Slivinski.
\newblock Theoria: Rewrite-acceptability verification over informal reasoning states.
\newblock \emph{arXiv:2607.01223}, 2026.

\bibitem[Geifman and El-Yaniv(2017)]{geifman2017selective}
Y.~Geifman and R.~El-Yaniv.
\newblock Selective classification for deep neural networks.
\newblock In \emph{NeurIPS}, 2017. arXiv:1705.08500.

\bibitem[Zhou et~al.(2026)]{zhou2026variation}
Y.~Zhou, A.~Xu, Y.~Zhou, J.~Singh, J.~Gui, and S.~Joty.
\newblock Variation in verification: Understanding verification dynamics in
  large language models.
\newblock In \emph{ICLR}, 2026.

\bibitem[Zheng et~al.(2023)]{zheng2023judging}
L.~Zheng et~al.
\newblock Judging {LLM}-as-a-judge with {MT-Bench} and {Chatbot Arena}.
\newblock In \emph{NeurIPS Datasets and Benchmarks}, 2023. arXiv:2306.05685.

\bibitem[Singhi et~al.(2025)]{singhi2025whentosolve}
N.~Singhi et~al.
\newblock When to solve, when to verify: Compute-optimal problem solving and
  generative verification for {LLM} reasoning.
\newblock In \emph{COLM}, 2025. arXiv:2504.01005.

\bibitem[Wang et~al.(2023)]{wang2023selfconsistency}
X.~Wang et~al.
\newblock Self-consistency improves chain of thought reasoning in language models.
\newblock In \emph{ICLR}, 2023. arXiv:2203.11171.

\bibitem[Ling et~al.(2023)]{ling2023deductive}
Z.~Ling et~al.
\newblock Deductive verification of chain-of-thought reasoning.
\newblock In \emph{NeurIPS}, 2023. arXiv:2306.03872.

\bibitem[Rein et~al.(2024)]{gpqa}
D.~Rein et~al.
\newblock {GPQA}: A graduate-level {Google}-proof {Q\&A} benchmark.
\newblock In \emph{COLM}, 2024. arXiv:2311.12022.

\bibitem[Balunovi\'c et~al.(2025)]{matharena}
M.~Balunovi\'c et~al.
\newblock {MathArena}: Evaluating {LLMs} on uncontaminated math competitions.
\newblock In \emph{NeurIPS Datasets and Benchmarks}, 2025. arXiv:2505.23281.

\bibitem[Pandit et~al.(2026)]{hard2verify}
S.~Pandit et~al.
\newblock {Hard2Verify}: A step-level verification benchmark for open-ended
  frontier math.
\newblock In \emph{ACL}, 2026. arXiv:2510.13744.

\bibitem[Hendrycks et~al.(2021)]{hendrycks2021math}
D.~Hendrycks et~al.
\newblock Measuring mathematical problem solving with the {MATH} dataset.
\newblock In \emph{NeurIPS Datasets and Benchmarks}, 2021. arXiv:2103.03874.

\bibitem[Lightman et~al.(2024)]{lightman2023lets}
H.~Lightman et~al.
\newblock Let's verify step by step.
\newblock In \emph{ICLR}, 2024. arXiv:2305.20050.

\bibitem[{Qwen Team}(2025)]{qwen3}
{Qwen Team}.
\newblock Qwen3 technical report.
\newblock \emph{arXiv:2505.09388}, 2025.

\bibitem[{OpenAI}(2025)]{gptoss}
{OpenAI}.
\newblock gpt-oss-120b \& gpt-oss-20b model card.
\newblock \emph{arXiv:2508.10925}, 2025.

\bibitem[Abdin et~al.(2024)]{phi4}
M.~Abdin et~al.
\newblock Phi-4 technical report.
\newblock \emph{arXiv:2412.08905}, 2024.

\bibitem[Norman et~al.(2026)]{norman2026reliability}
J.~D. Norman, M.~U. Rivera, and D.~A. Hughes.
\newblock Reliability without validity: A systematic, large-scale evaluation of
  {LLM}-as-a-judge models across agreement, consistency, and bias.
\newblock \emph{arXiv:2606.19544}, 2026.

\bibitem[Wu(2026)]{wu2026narratives}
W.~Wu.
\newblock When errors become narratives: A longitudinal taxonomy of silent
  failures in a production {LLM} agent runtime.
\newblock \emph{arXiv:2606.14589}, 2026.

\bibitem[Li et~al.(2026)]{preferenceleakage}
D.~Li et~al.
\newblock Preference leakage: A contamination problem in {LLM}-as-a-judge.
\newblock In \emph{ICLR}, 2026. arXiv:2502.01534.

\end{thebibliography}
\end{document}